\documentclass[sigconf,nonacm]{acmart}

\AtBeginDocument{%
  }

\setcopyright{none}
\copyrightyear{2025}
\acmYear{2025}
\usepackage{subcaption}
\usepackage{graphicx} 

\begin{document}

\title{Gradients: When Markets Meet Fine-tuning - A Distributed Approach to Model Optimisation}


\author{Christopher Subia-Waud}
\affiliation{%
  \institution{Rayon Labs}
  \country{UK} 
}
\email{ww@rayonlabs.ai}

\begin{abstract}
Current AutoML platforms leave substantial performance untapped. Testing 180 fine-tuning tasks across models from 70M to 70B parameters, we found that HuggingFace AutoTrain, TogetherAI, Databricks, and Google Cloud consistently produce suboptimal configurations. Gradients, built on the Bittensor network, attacks this problem through competition. Independent miners race to find optimal hyperparameters, earning rewards proportional to their models' performance. This tournament drives exploration of configuration spaces that single-strategy methods never examine. In our experiments, Gradients achieved a 100\% win rate against TogetherAI, Databricks, and Google Cloud, and beat HuggingFace AutoTrain in 82.8\% of experiments. Mean improvements reached 42.1\% against commercial platforms. Retrieval-augmented generation tasks saw 30-40\% gains; diffusion models improved 23.4\% on person-specific generation. When miners compete for rewards, they develop optimization strategies that centralized approaches overlook. These findings demonstrate that decentralized systems with economic incentives can systematically outperform traditional AutoML, suggesting market dynamics may be key to achieving superior fine-tuning results. Code is available at \url{https://github.com/rayonlabs/G.O.D}.
\end{abstract}


\begin{teaserfigure}
  \includegraphics[width=\textwidth]{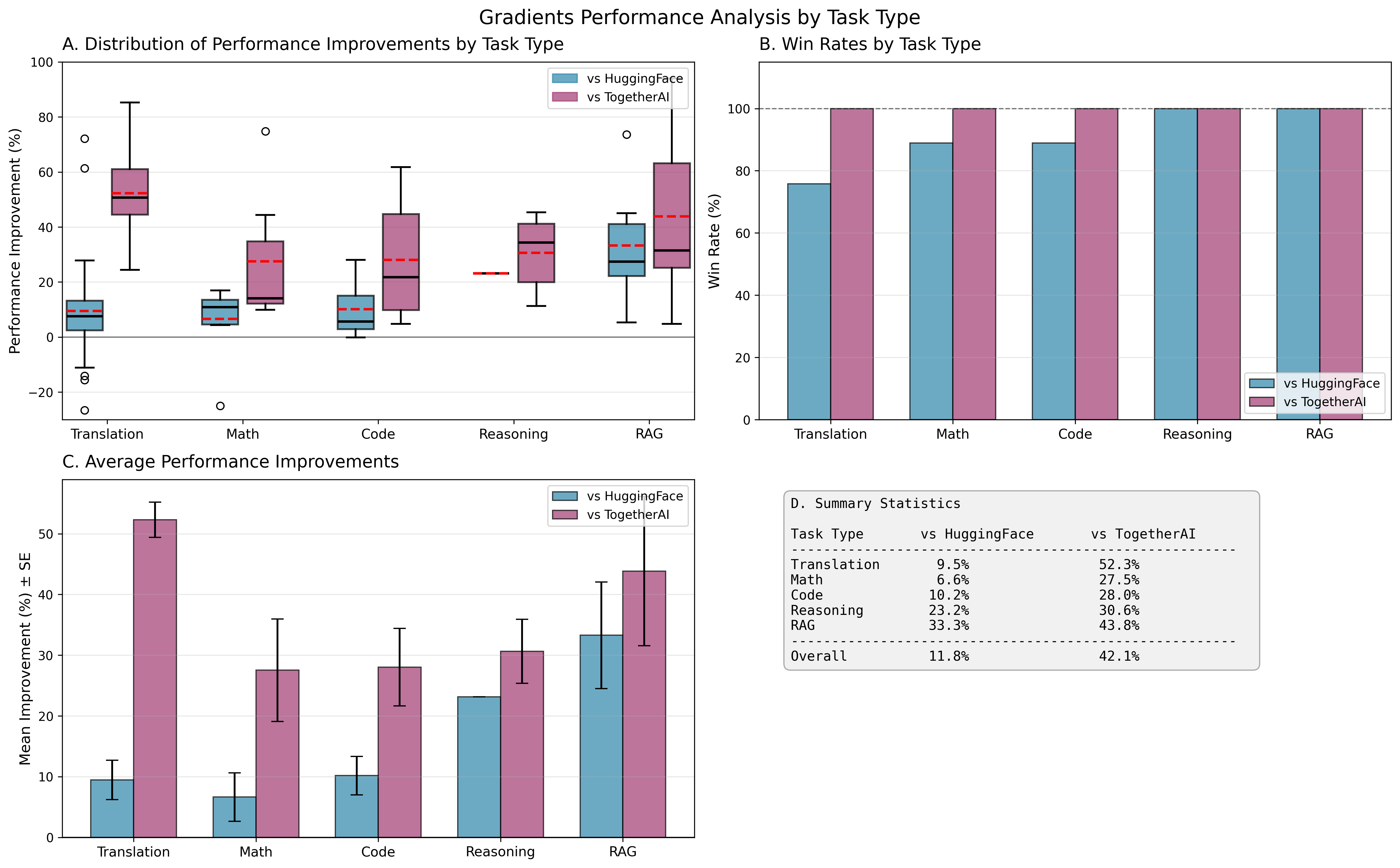}
  \caption{Performance improvement by task type. Four-panel analysis: box plots show improvement distributions, bar charts display win rates and mean improvements with standard errors, summary table provides exact values.}
  \label{fig:task_analysis}
\end{teaserfigure}

\maketitle

\section{The open challenges in model fine-tuning}

Fine-tuning foundation models has become a critical component of modern AI deployment, yet current AutoML platforms consistently produce suboptimal configurations. Our evaluation across 180 fine-tuning tasks reveals that HuggingFace AutoTrain, TogetherAI, Databricks, and Google Cloud leave performance gaps of 20-40\% in retrieval and reasoning tasks, suggesting that current approaches leave substantial performance unrealized.

The root cause lies in the nature of the fine-tuning problem itself. Fine-tuning is not a single optimization challenge but thousands of distinct problems~\cite{zhang2024scaling}. A 7B parameter model learning to generate code requires fundamentally different hyperparameters than the same model learning Japanese translation. A 70M model trained on 10,000 examples for 2 hours needs different settings than a 70B model trained on 500,000 examples for 10 hours. Current AutoML platforms~\cite{thakur2024autotrain,feurer2015efficient} address this diversity with single optimization strategies—whether Bayesian optimization, grid search, or evolutionary algorithms. These methods explore only a small fraction of the viable configuration space. They cannot adapt to the specific characteristics of each task, model, and dataset combination. The result is systematic underperformance: acceptable results that miss superior configurations.

Mechanism design and distributed optimization research offer an alternative approach~\cite{duetting2019optimal,ding2020incentive}. When solution spaces are vast and optimal strategies unknown, competitive mechanisms often outperform centralized planning~\cite{mcmahan2017communication,harris2019toward}. Consider how Linux development leverages thousands of independent contributors, each pursuing different improvements. Or how Bitcoin mining incentivizes global competition to secure the network. In machine learning specifically, population-based training~\cite{jaderberg2017population} suggests that competition between agents often lead to better solutions than single-agent optimization. These examples share a common principle: properly aligned incentives drive exploration of solution spaces that centralized approaches cannot effectively navigate.

Gradients implements this principle for foundation model fine-tuning. Built on the Bittensor network~\cite{rao2021bittensor}, it creates a marketplace where independent miners compete to find optimal configurations. When a fine-tuning task arrives, the system distributes it to multiple miners—typically 10-20 for language models, 15-25 for diffusion models. Each miner independently determines their approach: learning rate schedules, batch sizes, optimization algorithms, LoRA configurations, data augmentation strategies. They train models in parallel, each exploring different regions of the hyperparameter space. The system evaluates all submissions on held-out test data and synthetic benchmarks, rewarding miners proportionally to their models' performance. This creates direct economic incentives for discovering superior configurations.

The empirical results demonstrate the effectiveness of this approach. Across 180 controlled experiments, Gradients achieved a 100\% win rate against TogetherAI, Databricks, and Google Cloud. It outperformed HuggingFace AutoTrain—widely considered the state-of-the-art in automated fine-tuning—in 82.8\% of direct comparisons. Mean improvements reached 11.8\% against HuggingFace and 42.1\% against commercial cloud platforms. Retrieval-augmented generation tasks showed gains of 30-40\%, while diffusion models improved 23.4\% on person-specific generation tasks. These improvements span models from 70M to 70B parameters and diverse task types including translation, code generation, mathematical reasoning, and image generation.

These findings have implications beyond immediate performance gains. As foundation models grow larger and tasks become more specialized, the configuration space expands exponentially. No single algorithm, regardless of sophistication, can adequately explore these vast spaces. Competitive markets, however, naturally parallelize exploration across diverse strategies. When miners profit from finding better configurations, they develop novel optimization techniques, share successful strategies within their organizations, and continuously refine their approaches. The system's knowledge grows through evolutionary pressure rather than algorithmic updates. This suggests a fundamental shift in how we approach AutoML: from better algorithms to better incentive structures.

This paper presents the design, implementation, and evaluation of the Gradients system. Section 2 reviews related work in hyperparameter optimization, distributed machine learning, and economic mechanism design. Section 3 describes the Gradients architecture, including task distribution, miner selection, evaluation protocols, and reward mechanisms. Section 4 details our experimental methodology for comparing against existing AutoML platforms. Section 5 presents comprehensive results across model scales and task types. Section 6 discusses why competitive approaches succeed and their implications for future AutoML systems. We conclude by considering the broader potential of economically-driven optimization in machine learning. The theoretical foundation of this approach combines distributed optimization~\cite{li2017hyperband}, empirical risk minimization, and tournament theory~\cite{moldovanu2001optimal}, establishing a framework where properly structured competition can efficiently navigate complex optimization landscapes that resist traditional methods.

\section{Related Work}

\subsection{Hyperparameter Optimisation and AutoML} 
The field of automated hyperparameter optimisation has progressed from simple grid and random search methods to sophisticated multi-fidelity approaches. Bayesian optimisation frameworks like Hyperband~\cite{li2017hyperband} and BOHB~\cite{falkner2018bohb} combine principled uncertainty estimation with early stopping mechanisms to efficiently navigate large hyperparameter spaces. These methods achieve computational efficiency through bandit-based resource allocation, but remain fundamentally centralised in their coordination mechanisms. Recent AutoML platforms including AutoKeras~\cite{jin2019autokeras} and Auto-Sklearn~\cite{feurer2015efficient} have democratised access to these techniques, yet they still rely on single optimisation strategies that may miss task-specific optima. HuggingFace AutoTrain~\cite{thakur2024autotrain} represents the current state-of-the-art in no-code model fine-tuning, providing automated hyperparameter optimisation across multiple modalities but using centralised search strategies.

\subsection{Distributed and Competitive Optimisation} 
Population-based training (PBT)~\cite{jaderberg2017population} introduced competitive dynamics to neural network optimisation, where populations of models compete and evolve hyperparameters during training. This approach demonstrates how competition discovers superior configurations through exploitation and exploration mechanisms reminiscent of evolutionary algorithms~\cite{dushatskiy2023multi}. Parallel to this, federated learning~\cite{mcmahan2017communication} established principles for coordinating distributed model training without centralising data, achieving communication efficiency through local computation and periodic aggregation. However, these approaches lack economic incentive mechanisms to align individual participant behaviour with collective optimisation goals.

\subsection{Parameter-Efficient Fine-tuning} 
The emergence of foundation models has necessitated new approaches to efficient adaptation. LoRA~\cite{hu2022lora} and its variants inject trainable low-rank matrices into pre-trained transformers, reducing trainable parameters by orders of magnitude whilst preserving performance. AdaLoRA~\cite{zhang2023adalora} extends this approach with adaptive rank allocation, whilst QLoRA~\cite{dettmers2023qlora} combines quantisation with low-rank adaptation for memory efficiency. These parameter-efficient methods are widely used by AutoML LLM and diffusion training platforms~\cite{ding2024parameter}.

\subsection{Economic Mechanisms in Machine Learning} 
Recent work explores how economic incentives coordinate machine learning systems. Differentiable economics~\cite{duetting2019optimal} employs neural networks to automatically design optimal auction mechanisms, demonstrating how learning systems embed economic principles. Blockchain-based coordination mechanisms~\cite{harris2019toward,shayan2021biscotti} show promise for incentivising distributed computation, whilst tournament theory provides mathematical foundations for competition-based optimisation~\cite{moldovanu2001optimal}. Incentive mechanism design for distributed coded machine learning~\cite{ding2020incentive} establishes optimal reward structures for heterogeneous computational contributors. However, researchers have largely left unexplored the application of these economic principles to automated hyperparameter optimisation.

\subsection{Decentralised Machine Learning Networks} 
Emerging platforms like Bittensor~\cite{rao2021bittensor} demonstrate how cryptoeconomic incentives coordinate distributed machine learning tasks. These networks employ token-based rewards to align individual computational contributions with network-wide objectives, creating sustainable ecosystems for collaborative AI development. The Yuma Consensus mechanism~\cite{bittensor2025dtao} provides robust peer-ranking systems resistant to collusion, whilst maintaining decentralised coordination. Our work builds on these foundations by specifically applying economic coordination mechanisms to the hyperparameter optimisation problem, demonstrating how competitive markets discover superior model configurations.

\subsection{Foundation Model Fine-tuning and Evaluation} 
Recent advances in fine-tuning large language models focus on instruction following~\cite{ouyang2022training}, preference optimisation~\cite{rafailov2024direct}, and synthetic data generation~\cite{chan2024synthetic}. For diffusion models, techniques like DreamBooth~\cite{ruiz2023dreambooth} enable subject-driven generation through personalised fine-tuning. Community-driven platforms like CivitAI~\cite{wei2024exploring} have democratised access to fine-tuned diffusion models, though require an understanding of the diffusion training process or the use of default parameters in order to use. Our work addresses this gap by providing economic incentives for discovering optimal fine-tuning configurations across diverse model types and tasks without the need for users of the platform to have a deep understanding of the fine-tuning process.

\section{Gradients}

The primary purpose of Gradients is to serve real users with high-quality fine-tuning solutions for practical applications. Unlike traditional AutoML systems that rely on a single optimisation strategy or limited search methods, our approach transforms fine-tuning into a competitive tournament where participants are rewarded based on the quality of their solutions relative to others. This approach enables exploration of a much broader solution space to better meet genuine user needs, while supplementary synthetic benchmark tasks help maintain system quality and miner skill development.

\subsection{System Architecture}

The Gradients system consists of three principal actors and multiple computational processes:

\noindent \textbf{Validators} are network nodes responsible for task creation, dataset preparation, and submission evaluation. They maintain the integrity of the system through fair scoring and reward distribution. Crucially, validators also channel organic tasks from real users into the network, ensuring the system solves practical fine-tuning challenges alongside synthetic benchmarks.

\noindent \textbf{Miners} are independent computational participants who compete to find optimal fine-tuning configurations. They receive task specifications and submit fine-tuned models for evaluation.

\noindent  \textbf{Tasks} represent specific fine-tuning challenges, each with defined datasets, model architectures, and evaluation criteria. Tasks may be organic (derived from real user needs) or synthetic (created for systematic benchmarking).

The system operates through a continuous cycle of task creation, miner assignment, model training, and evaluation, with economic rewards distributed according to performance.

The core idea of Gradients lies in its transformation of hyperparameter optimisation into a competitive marketplace, creating three key advantages. First, independent miners employ diverse approaches to the same optimisation problem, ensuring strategy diversity across the network. Second, multiple regions of hyperparameter space are explored simultaneously through parallel exploration, maximising coverage of potential solutions. Third, competitive rewards drive continuous refinement of optimisation techniques through evolutionary pressure, naturally selecting for the most effective approaches. Miners operate as independent agents with complete freedom in determining their fine-tuning approach, including choices in hyperparameter selection such as learning rates and batch sizes, optimisation techniques like LoRA, QLoRA, or full fine-tuning, hardware allocation and parallelisation strategies, and custom augmentation or preprocessing methods. This autonomy creates natural strategy diversity that evolves through competitive pressure, driving innovation without requiring centralised coordination and ensuring that the most effective optimisation strategies emerge organically through market dynamics.

\subsection{Decentralised Task Distribution}

The task distribution mechanism ensures fair allocation of fine-tuning opportunities while maintaining incentives for high-quality submissions. Validators create tasks by specifying $\mathcal{T} = \langle M, D, \Delta t, \mathcal{E} \rangle$ where $M$ represents the base model identifier, $D$ the dataset specifications, $\Delta t$ the allocated time for completion, and $\mathcal{E}$ the evaluation criteria.

The system prioritises organic tasks—those created from real user needs—as these represent the actual problems people need solved. These organic tasks are the primary purpose of the Gradients system and provide real-world validation of fine-tuning techniques. To supplement the organic workload and maintain consistent network activity, the system also creates synthetic tasks through a controlled distribution mechanism. Synthetic tasks provide consistent benchmarks for miner skill development, ensure comprehensive coverage of model types and domains, and maintain network activity during periods of lower organic demand.

Tasks fall into four technical categories:

\noindent  \textbf{Instruction Tuning Tasks} for training language models on instruction-following datasets.

\noindent \textbf{Direct Preference Optimisation (DPO) Tasks} for preference-based optimisation of language models.

\noindent \textbf{Group Relative Policy Optimisation (GRPO) Tasks} for advanced preference-based training.

\noindent \textbf{Image Generation Tasks} for fine-tuning diffusion models.

Tasks are created with a controlled distribution across categories to ensure balanced coverage:

\begin{equation}
P(\text{TaskType} = t) =
\begin{cases}
\rho_{\text{instruct}} & \text{if } t = \text{InstructTextTask} \\
\rho_{\text{dpo}} & \text{if } t = \text{DPOTask} \\
\rho_{\text{grpo}} & \text{if } t = \text{GRPOTask} \\
\rho_{\text{image}} & \text{if } t = \text{ImageTask}
\end{cases}
\end{equation}

with parameters $\rho_{\text{instruct}} = 0.25$, $\rho_{\text{dpo}} = 0.1$, $\rho_{\text{grpo}} = 0.3$, and $\rho_{\text{image}} = 0.35$, at the time of writing this with these weightings being adjusted to complement the organic job proportions.

For each task, the system selects a pool of miners through a weighted probability mechanism designed to balance exploration and exploitation. Given a set of available miners $\mathcal{M} = \{m_1, m_2, ..., m_n\}$, selection weights are assigned as $w_i = \alpha$ if miner $m_i$ has not participated today, or $w_i = \max(s_i, \gamma)$ otherwise, where $\alpha$ is the default score for first daily participation (set to 2.0), $s_i$ is miner $m_i$'s average quality score from previous tasks, and $\gamma$ is the minimum score threshold (0.01).

Miners are then sorted by weight and assigned position-based selection probabilities:
\begin{equation}
p_i = \lambda - (i-1) \cdot \frac{\lambda - 1}{|\mathcal{M}|}
\end{equation}
where $p_i$ is the relative probability of selecting the $i$-th ranked miner, $\lambda$ is the top miner chance multiplier (3.0), and $|\mathcal{M}|$ is the total number of available miners.

This approach preferentially selects high-performing miners while ensuring all miners maintain participation opportunities, preventing monopolisation by established participants.

\subsection{Dataset Preparation Pipeline}

Effective dataset preparation ensures fair evaluation and meaningful comparison of fine-tuning approaches. For each task, the system partitions the available data into three distinct sets: $D = D_{\text{train}} \cup D_{\text{test}} \cup D_{\text{synth}}$ where $D_{\text{train}}$ is provided to miners for model fine-tuning, $D_{\text{test}}$ is used for primary performance evaluation with size $|D_{\text{test}}| = \min(|D| \cdot \rho_{\text{test}}, \kappa_{\text{test}})$, and $D_{\text{synth}}$ is algorithmically generated to test generalization with size $|D_{\text{synth}}| = \min(|D| \cdot \rho_{\text{synth}}, \kappa_{\text{synth}})$, using parameters $\rho_{\text{test}} = 0.1$, $\rho_{\text{synth}} = 1.0$, $\kappa_{\text{test}} = 1,000$, and $\kappa_{\text{synth}} = 300$.

Synthetic data generation serves multiple purposes: it provides an additional evaluation benchmark, tests for generalisation capabilities, and mitigates overfitting to the test set. For language models, using a controlled process with temperature $\tau = 0.6$, we generate synthetic data that preserves schema requirements while introducing semantic diversity through the process $x'_i = G(x_i, \tau)$ for all $x_i \in D_{\text{sample}}$, where $G$ is a high-quality foundation model generating function. For DPO tasks, this includes generating both chosen and rejected responses using models of different capabilities. For diffusion models, using temperature $\tau = 0.4$, the system generates new image-text pairs $(I_i, P_i) = G_{\text{image}}(S_i, R_i)$ with controlled style variations and prompt diversity, where $S_i$ is a style specification and $R_i$ is a target resolution. Image resolutions are standardised to ensure fair comparison, with dimensions constrained to be divisible by 64.

\subsubsection{Task Time Allocation}

The allocated time for task completion scales with both dataset size and model complexity:
\begin{equation}
\Delta t \in [f_{\min}(|D|, |M|), f_{\max}(|D|, |M|)]
\end{equation}

For text tasks, time allocation follows a binned approach:
\begin{equation}
\Delta t \in
\begin{cases}
[3, 6] \text{ hours} & \text{if } |D| \in [10000, 25000] \\
[4, 8] \text{ hours} & \text{if } |D| \in [25000, 50000] \\
[5, 9] \text{ hours} & \text{if } |D| \in [50000, 100000] \\
[7, 10] \text{ hours} & \text{if } |D| \in [100000, 500000]
\end{cases}
\end{equation}

For image tasks, time allocation is generally shorter due to the efficiency of LoRA-based fine-tuning for diffusion models, typically ranging from 1-2 hours regardless of dataset size.

\subsection{Multi-Level Evaluation Framework}

To ensure fair comparison and prevent gaming, all models are evaluated in isolated Docker environments with standardised GPU configurations. This creates a controlled evaluation sandbox where performance differences reflect genuine optimisation quality rather than environmental variations.

For language models, the evaluation process computes both test and synthetic losses:
\begin{align}
L_{\text{test}}(m_i) &= \frac{1}{|D_{\text{test}}|}\sum_{x \in D_{\text{test}}} \ell(M_{m_i}, x) \\
L_{\text{synth}}(m_i) &= \frac{1}{|D_{\text{synth}}|}\sum_{x \in D_{\text{synth}}} \ell(M_{m_i}, x)
\end{align}
where $M_{m_i}$ is the model submitted by miner $m_i$ and $\ell$ is an appropriate loss function.

For diffusion models, evaluation encompasses text-guided and non-text-guided image reconstruction quality:
\begin{align}
L_{\text{text}}(m_i) &= \frac{1}{|D_{\text{test}}|}\sum_{(I,p) \in D_{\text{test}}} \|I - M_{m_i}(I_{\text{noisy}}, p)\|_2^2 \\
L_{\text{no-text}}(m_i) &= \frac{1}{|D_{\text{test}}|}\sum_{I \in D_{\text{test}}} \|I - M_{m_i}(I_{\text{noisy}}, \emptyset)\|_2^2
\end{align}
where $I$ is the original target image, $I_{\text{noisy}}$ is the noise-corrupted input, and $p$ is the text prompt.

To prevent overfitting to either evaluation set, performance is measured as a weighted combination $L_{\text{weighted}}(m_i) = \omega \cdot L_{\text{test}}(m_i) + (1-\omega) \cdot L_{\text{synth}}(m_i)$ where $\omega$ is the test score weighting parameter (0.7 for text tasks). For image models, the weighting combines text-guided and non-text-guided metrics: $L_{\text{weighted}}(m_i) = \delta \cdot L_{\text{text}}(m_i) + (1-\delta) \cdot L_{\text{no-text}}(m_i)$ with $\delta = 0.25$ emphasising coherence over prompt adherence.

\subsection{Miner Scoring}

The scoring system translates raw performance metrics into reward signals that drive system behaviour, using a multi-level approach that balances immediate task performance with long-term incentive alignment. Within each task, miners are ranked by their weighted loss, with scores assigned according to:
\begin{equation}
S_{\text{task}}(m_i) =
\begin{cases}
S_{\text{first}} & \text{if rank}(m_i) = 1 \\
S_{\text{penalty}} & \text{if rank}(m_i) > |\mathcal{M}_{\text{valid}}| \cdot (1 - \rho_{\text{penalty}}) \\
0 & \text{otherwise}
\end{cases}
\end{equation}

with parameters $S_{\text{first}} = 3.0$ (first-place score) and $S_{\text{penalty}} = -1.0$ (penalty score). The bottom 25\% of miners by performance receive the penalty score. This creates strong incentives for top performance while discouraging poor-quality submissions that waste network resources.

Raw scores are adjusted based on task complexity to ensure fair comparison across different fine-tuning challenges:
\begin{equation}
S_{\text{adjusted}}(m_i, \mathcal{T}) = S_{\text{task}}(m_i) \cdot W_{\text{task}}(\mathcal{T})
\end{equation}
where the task weight function $W_{\text{task}}(\mathcal{T}) = \max(1, 2 \cdot \sqrt{\phi(M) \cdot t})$ incorporates model size ($\phi(M)$ in billions of parameters) and training time allocation ($t$ in hours).

Performance is aggregated across multiple time windows to balance recent achievements with established track records:
\begin{equation}
S_{\text{temporal}}(m_i) = \sum_{w \in W} \omega_w \cdot \text{Normalize}\left(\sum_{\mathcal{T} \in \mathcal{T}w} S{\text{adjusted}}(m_i, \mathcal{T})\right)
\end{equation}
where $W$ is the set of time windows \{1-day, 3-day, 7-day\}, $\omega_w$ is the weight for window $w$ (0.3, 0.3, and 0.4 respectively), and $\mathcal{T}_w$ is the set of tasks in window $w$.

The final score undergoes a non-linear transformation to create appropriate reward differentials:
\begin{equation}
S_{\text{final}}(m_i) = \beta \cdot S_{\text{sigmoid}}(m_i) + \omega \cdot x_i
\end{equation}
where:
\begin{align}
S_{\text{sigmoid}}(m_i) &= \left(\frac{1}{1 + \exp(-\gamma \cdot (x_i - \mu))}\right)^{\nu} 
\end{align}
and $x_i = \frac{\text{quality\_score}_i}{\text{max\_score}}$ is the normalised input score, with parameters $\beta = 0.7$ (sigmoid weight), $\omega = 0.05$ (linear weight), $\gamma = 9$ (sigmoid steepness), $\mu = 0.5$ (sigmoid shift), and $\nu = 0.75$ (sigmoid power).

This transformation creates pronounced rewards for top performers while maintaining reasonable incentives for all participants.

\subsection{Blockchain Integration and Economic Incentives}

The final scores are translated into on-chain weights that determine token emissions from the underlying Bittensor network: $W_{\text{chain}}(m_i) = S_{\text{final}}(m_i) \cdot \text{Vtrust}$.

\subsection{System Fault Tolerance}

Gradients additionally provides inherent fault tolerance through several complementary mechanisms. When insufficient miners accept a task, the system enters a delay state and reattempts the task later with extended time allocation following $\Delta t_{\text{delay}} = \Delta t_{\text{initial}} + c \cdot \text{attempts}$. The assignment of multiple miners to each task (8-15 for text tasks, 15-25 for image tasks) creates redundancy that ensures successful completion even when individual miners encounter hardware failures or connectivity issues. Additionally, the weighted selection process naturally evolves to favour more reliable participants, as miners with higher success rates receive increased selection probabilities over time. These mechanisms work together to maintain system operability despite the inherent unpredictability of distributed computational environments, including fluctuations in miner participation, hardware failures, and network connectivity problems.

Having described the system design and competitive mechanism, we now evaluate whether this approach actually produces superior results in practice. Our experimental methodology tests this across diverse foundation models and task types.

\section{Methodology}

\subsection{Experimental Design}

We conducted 180 controlled experiments to evaluate the effectiveness of decentralised hyperparameter optimisation across diverse models, datasets, and task types. To ensure fair comparison, all experiments used identical conditions across platforms: the same base models, datasets, training time allocations, and evaluation metrics. We trained all models using the Gradients platform's distributed fine-tuning approach, then evaluated them using task-specific metrics against baseline implementations from leading AutoML services using these identical experimental conditions. We focus the experimental space on instruct and diffusion tasks - since there are no other platforms to reliably compare DPO and GRPO tasks.

\subsection{Model Selection}

We tested five model families spanning three orders of magnitude in size. EleutherAI Pythia-70M~\cite{biderman2023pythia} represents the smallest scale, testing extreme parameter efficiency. The Qwen2 family~\cite{yang2024qwen2} (1.5B, 7B) provides popular open models with strong multilingual capabilities. Llama-3.1-8B and Llama-3.3-70B~\cite{grattafiori2024llama3,llama33model2024} are Meta's latest instruction-tuned models. DeepSeek-R1-Distill-70B~\cite{deepseekr12025} represents state-of-the-art reasoning-optimised architecture.

For diffusion models, we evaluated two state-of-the-art text-to-image architectures: SDXL 1.0~\cite{podell2023sdxl} (Stable Diffusion XL), the workhorse of open image generation, and Flux.1~\cite{flux2024}, a next-generation architecture with improved coherence.

\subsection{Training Protocol}

We trained all models using the Gradients decentralised platform as described in Section 3. The training process used competitive hyperparameter optimisation where validators prepared train/test/synthetic splits (typically 80/10/synth), the system selected 8-15 miners for text tasks and 15-25 for image tasks via weighted probability based on historical performance, the platform allocated time based on dataset size (3-10 hours for text, 1-2 hours for images), and miners exercised complete autonomy in selecting optimisation strategies, learning rates, batch sizes, and training techniques.

\subsection{Evaluation Methodology}

\subsubsection{Language Model Evaluation}

For text models, we computed direct loss on held-out test sets. We calculated the cross-entropy loss on the test split using the fine-tuned model, applied no additional post-processing or ensemble methods, and results represent the raw predictive performance of each fine-tuning approach.

\subsubsection{Diffusion Model Evaluation}

For diffusion models, we implemented a dual evaluation protocol that tests the model's ability to reconstruct original images through a controlled denoising process:

\noindent \textbf{Text-Guided Generation:} Models received test images with added noise and accompanying text prompts, then attempted to reconstruct the original images. We measured reconstruction quality via L2 pixel loss:
\begin{equation}
L_{\text{text}} = \frac{1}{N}\sum_{i=1}^{N} \|I_{\text{original},i} - G(I_{\text{noisy},i}, p_i)\|_2^2
\end{equation}
where $G$ is the fine-tuned generator, $I_{\text{noisy},i}$ is the noise-corrupted input image, $p_i$ is the text prompt, and $I_{\text{original},i}$ is the target original image.

\noindent \textbf{Non-Text-Guided Generation:} Models performed the same image reconstruction task without text conditioning to evaluate learned style consistency:
\begin{equation}
L_{\text{no-text}} = \frac{1}{N}\sum_{i=1}^{N} \|I_{\text{original},i} - G(I_{\text{noisy},i}, \emptyset)\|_2^2
\end{equation}

We evaluated each model at 9 noise levels (0.1-0.9 denoise strength) with 10 seeds per level, testing robustness across denoising strengths.

\subsection{Baseline Comparisons}

We benchmarked Gradients against four leading AutoML services:

\noindent \textbf{HuggingFace AutoTrain}~\cite{thakur2024autotrain}: State-of-the-art AutoML with Bayesian hyperparameter optimisation

\noindent \textbf{TogetherAI Fine-tuning}: Cloud platform using standard fine-tuning configurations

\noindent \textbf{Databricks}: Enterprise MLOps platform (limited experiments due to the platform removing fine-tuning options part-way through our experiments)

\noindent \textbf{Google Cloud Vertex AI}: GCP's managed AutoML service (limited experiments due to prohibitive costs - \$10,000+ for a single 70B model training run)

For diffusion models, we compared against CivitAI~\cite{wei2024exploring}, the only competitor available for this form of lora training.

\subsection{Language Model Tasks}

We evaluated language model fine-tuning across five categories representing diverse reasoning capabilities and linguistic complexity. These tasks span fundamental NLP challenges from mathematical reasoning to multilingual translation, testing the platform's ability to optimise hyperparameters across varied domains. Each task category presents distinct optimisation challenges due to differences in data structure, reasoning requirements, and evaluation metrics.

\noindent \textbf{Mathematical Reasoning:} GSM8K~\cite{cobbe2021gsm8k} (grade school maths), OpenThoughts-114k~\cite{openthoughts2025} (step-by-step solutions), and verified mathematical solutions with chain-of-thought reasoning.

\noindent \textbf{Translation:} Eight language pairs from Opus-100~\cite{zhang2020opus100} and WMT19~\cite{wmt2019}: English$\leftrightarrow$\{French, Spanish, Japanese, Korean, Portuguese, Turkish, Russian, Chinese\}, plus challenging low-resource pairs (Lithuanian$\rightarrow$English, Kazakh$\rightarrow$English, Gujarati$\rightarrow$English).

\noindent \textbf{Code Generation:} CodeAlpaca-20k~\cite{codealpaca} (instruction-following), self-instruct-starcoder~\cite{selfinstructstarcoder2023} (compilation tasks), and CodeFeedback-Filtered (complex programming challenges).

\noindent \textbf{Retrieval-Augmented Generation:} RAG-Instruct~\cite{raginstruct2024} (document-grounded QA), Ruler~\cite{hsieh2024ruler} (long-context retrieval), and RAGBench~\cite{tang2024multihoprag,friel2024ragbench} (multi-hop reasoning).

\noindent \textbf{Reasoning:} Codeforces chain-of-thought, medical reasoning with explanations, and curated reasoning collections requiring multi-step inference.

\subsection{Diffusion Model Tasks}

For diffusion model evaluation, we focused on two primary challenge categories that test different aspects of fine-tuning effectiveness. These tasks assess the model's ability to learn specific visual concepts and artistic styles while maintaining generation quality and coherence. The task categories represent common real-world applications where users seek to customise diffusion models for specific aesthetic or identity requirements.

\noindent \textbf{Person-Specific:} 16 individuals with 10-50 images each, testing identity preservation across poses, lighting, and contexts.

\noindent \textbf{Style Transfer:} 18 artistic styles including cyberpunk, neoclassical, romanticism, constructivism, pointillism, pop art, and various cultural aesthetics (ukiyo-e, vintage anime).

\section{Results}

Our experimental evaluation tests whether economic incentives drive miners to discover task-specific optimisation strategies that consistently outperform standard approaches. The results support this hypothesis: competitive pressure leads to systematic performance improvements across diverse problem types.

The effectiveness of decentralised optimisation varies by task complexity. Figure~\ref{fig:task_analysis} shows  breakdown across task types. RAG and reasoning tasks—requiring complex contextual understanding—show the largest gains, while math show more modest, but consistent improvements.

\begin{figure}[htbp]
\centering
\includegraphics[width=0.8\columnwidth]{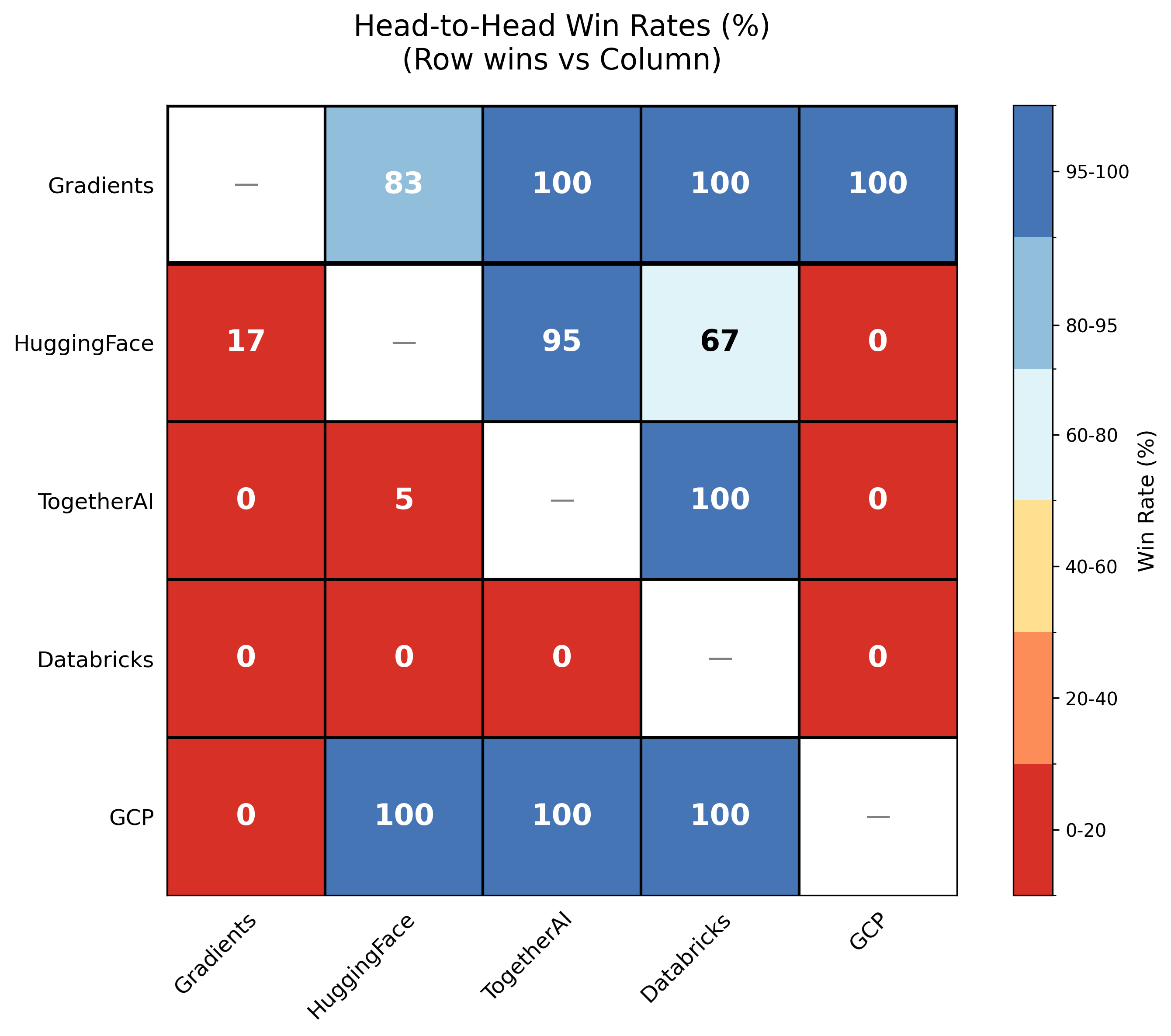}
\caption{Win rates against leading AutoML platforms. Matrix shows percentage of experiments where Gradients achieved lower loss than competitors across 180 controlled experiments.}
\label{fig:win_rates}
\end{figure}

Figure~\ref{fig:win_rates} demonstrates consistent superiority across all competitors, achieving 82.8\% win rate against HuggingFace AutoTrain and 100\% against TogetherAI, Databricks, and Google Cloud across 180 controlled experiments.

\begin{figure}[htbp]
\centering
\begin{subfigure}[b]{0.8\columnwidth}
\centering
\includegraphics[width=\linewidth]{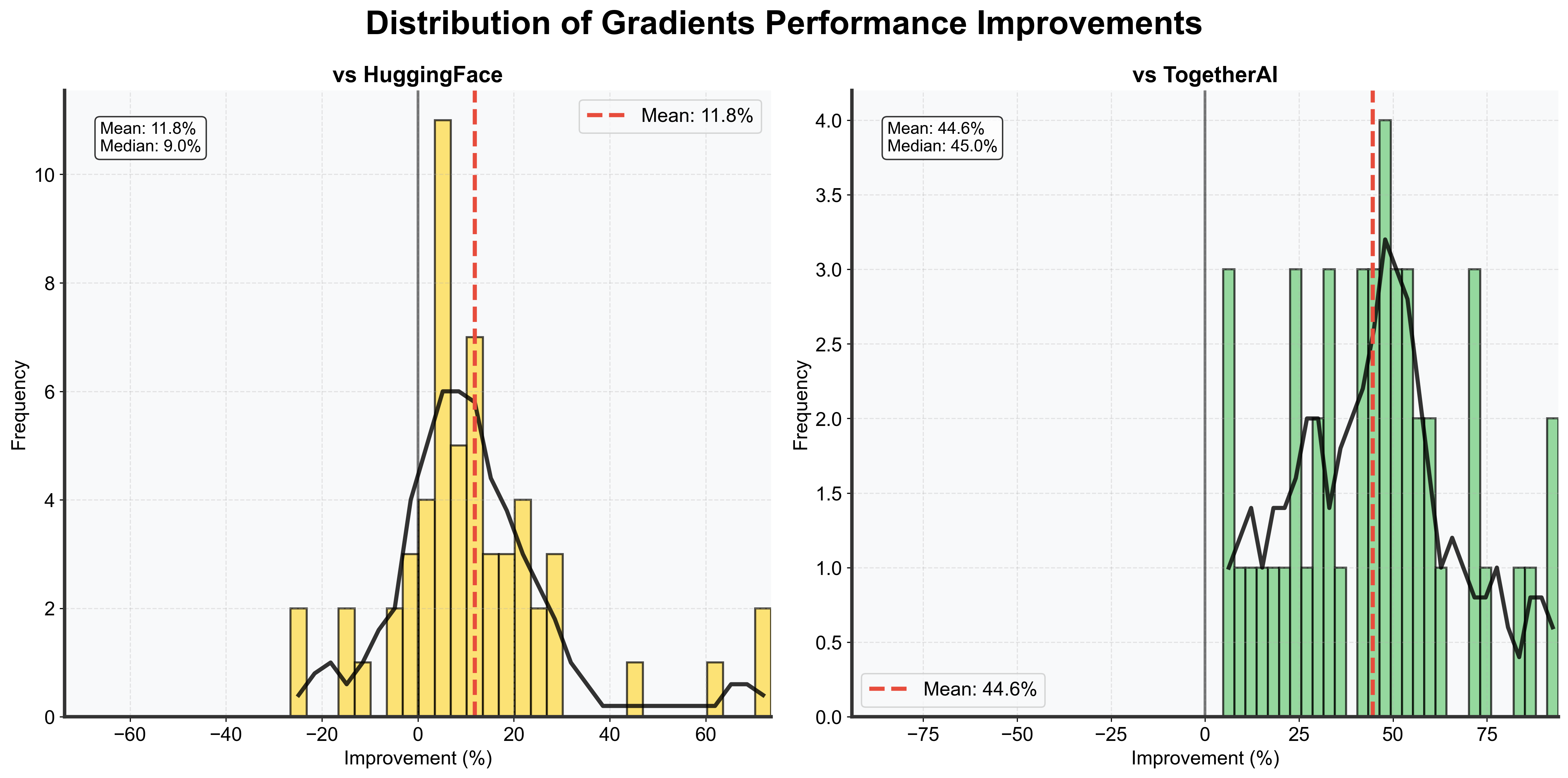}
\caption{Improvement distributions across experiments}
\label{fig:improvement_dists}
\end{subfigure}

\vspace{0.5cm}

\begin{subfigure}[b]{0.8\columnwidth}
\centering
\includegraphics[width=\linewidth]{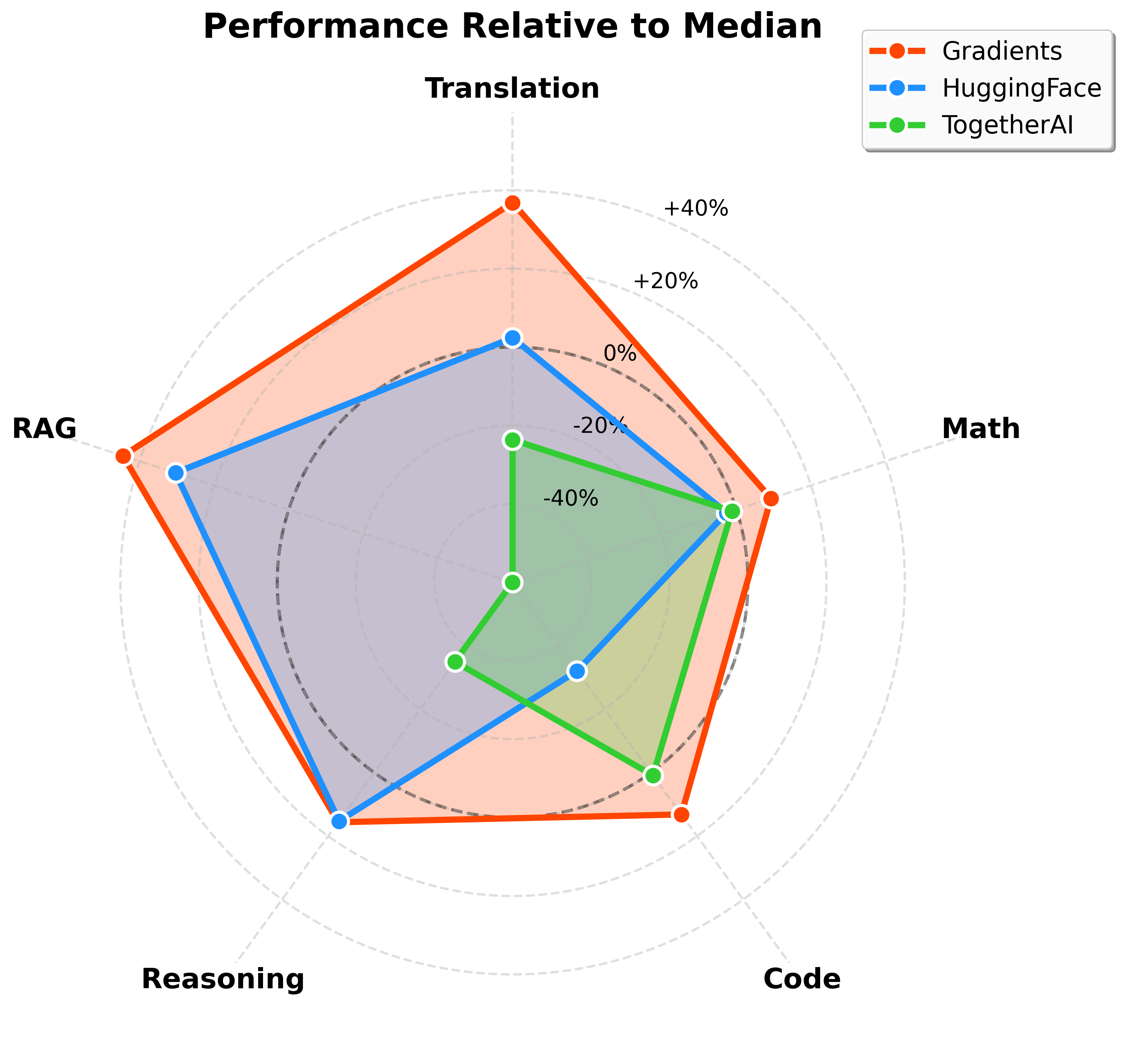}
\caption{Performance relative to median}
\label{fig:spider}
\end{subfigure}
\caption{Performance magnitude analysis. (a) Improvement distributions across all experiments demonstrate both consistency and magnitude of gains over competitors. (b) Spider chart visualises performance relative to cross-provider median, where positive values indicate above-median performance.}
\label{fig:overall_analysis}
\end{figure}

Figure~\ref{fig:overall_analysis} reveals the magnitude of these improvements, with distributions showing substantial gains—averaging 11.8\% against HuggingFace and 42.1\% against TogetherAI.

\subsection{Model Scale Effects}

\begin{figure}[htbp]
\centering
\begin{subfigure}[b]{0.8\columnwidth}
\centering
\includegraphics[width=\linewidth]{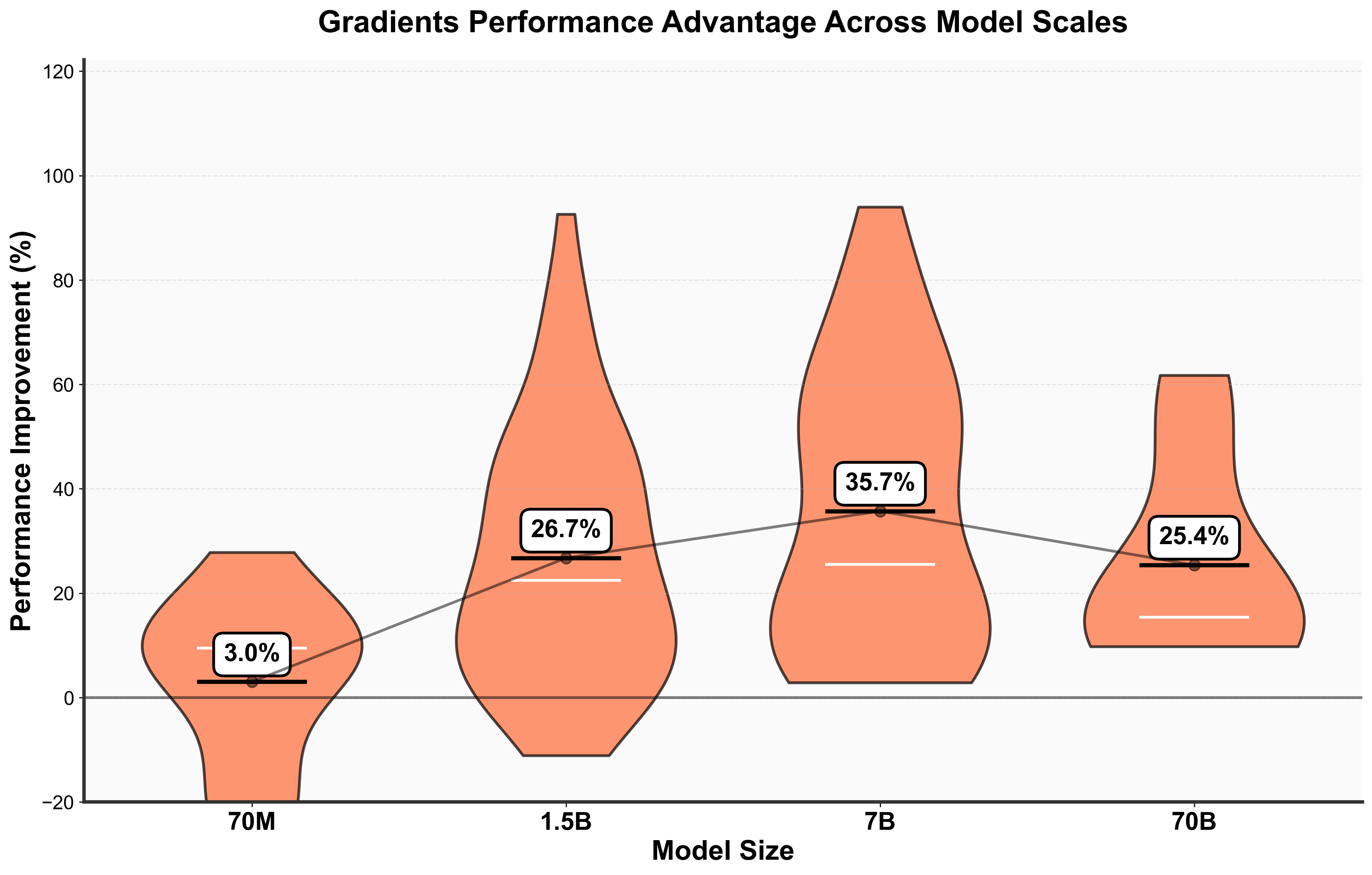}
\caption{Language model scale effects}
\label{fig:scale_effects}
\end{subfigure}

\vspace{0.5cm} 

\begin{subfigure}[b]{0.8\columnwidth}
\centering
\includegraphics[width=\linewidth]{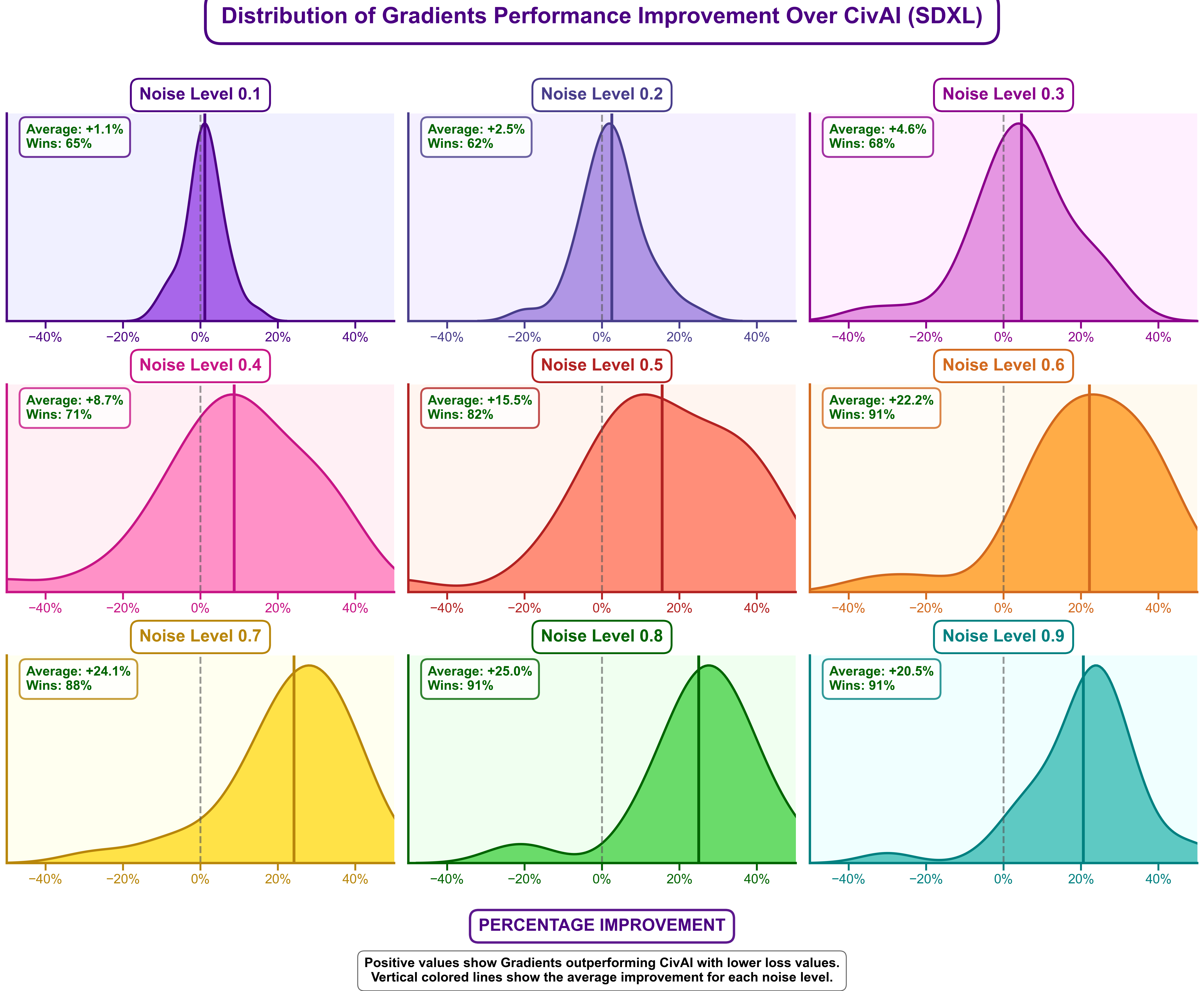}
\caption{Diffusion model improvements}
\label{fig:diffusion_sub}
\end{subfigure}
\caption{Scale and architecture effects. (a) Performance improvements peak at 7-8B parameters, suggesting optimal hyperparameter sensitivity. Very small (70M) and large (70B) models show reduced gains. (b) Diffusion model improvements increase with task difficulty - Gradients' advantage grows substantially as noise levels rise, demonstrating superior optimization for challenging denoising scenarios.}
\label{fig:scale_diffusion}
\end{figure}

Figure~\ref{fig:scale_diffusion} shows the relationship between model scale and performance improvements. We observe that mid-scale models (7-8B parameters) achieve the largest gains from competitive hyperparameter optimisation, while both smaller (70M) and larger (70B) models show more modest improvements. For diffusion models, we evaluated performance across varying noise levels (0.1 to 0.9 denoise strength). The competitive advantage of Gradients increases as the denoising task becomes more challenging—the performance gap widens substantially at higher noise levels, demonstrating that the distributed approach excels particularly when the reconstruction task is most difficult. Note that the number of experiments for 70B models was limited due to computational costs (exceeding \$10,000 per training run on some platforms), which may affect the statistical significance of results at that scale.

\section{Discussion}
  
\noindent \textbf{Why Competition Discovers Better Configurations:} Our results suggest three mechanisms drive competitive advantage in hyperparameter optimization. First, centralized search explores limited regions of configuration space—the consistency of our wins indicates that standard AutoML confines itself to local optima while miners discover superior global configurations. Second, the tournament structure fundamentally alters exploration incentives: miners must outperform competitors rather than achieve merely acceptable performance, naturally driving search into unexplored regions. Third, strategy opacity prevents convergence—because miners keep configurations proprietary, the system maintains diversity rather than collapsing to common approaches.

\noindent \textbf{Adaptation Without Central Coordination:} Traditional AutoML platforms improve through explicit algorithmic updates—a slow cycle from research to deployment. Gradients evolves continuously through market dynamics. During our experimental period, we observed progressive sophistication: early submissions used standard techniques while later submissions incorporated adaptive schedules and novel lora techniques. This distributed innovation emerges from economic pressure alone. 

\noindent \textbf{Computational Trade-offs:}

The primary limitation of competitive optimization is resource consumption. Each task requires 8-25 parallel training runs versus single runs in centralized systems, proportionally increasing energy use and cost. Whether 20-40\% performance gains justify this overhead depends on application requirements. High-stakes deployments may warrant the expense, while routine fine-tuning may not. Future work should explore adaptive competition—varying miner pools based on task novelty and importance—to balance discovery with efficiency.

\noindent \textbf{Strategic Opacity and Emergent Optimization:}

Unlike traditional AutoML where optimization paths are internally observable, Gradients operates as a black box we observe superior performance without visibility into
  specific winning configurations. This opacity is not a design flaw but an essential feature that drives discovery. Strategy secrecy preserves competitive advantage,
  incentivizing miners to explore novel combinations rather than converge on known parameters. This mirrors biological evolution and financial markets, where complex
  adaptive behavior emerges without central coordination.

  Despite configuration opacity, output artifacts reveal strategic patterns that contradict theoretical guidance. Miners consistently employ rsLoRA with ranks of
  64-128, despite theory suggesting rank 1 should suffice~\cite{hu2022lora}. They select rsLoRA for small datasets but prefer full fine-tuning for larger
  corpuses—empirical discoveries that conventional AutoML might never explore due to theoretical priors. This creates a paradox: the system's effectiveness may depend
  on its inscrutability. Traditional AutoML achieves reproducibility through transparency but limits search to theoretically justified strategies. Gradients discovers
  superior configurations precisely because miners explore regions that gradient-based or Bayesian optimization would never sample—using ranks orders of magnitude
  higher than theory suggests or learning rates far outside conventional ranges.

  Accepting emergent complexity may be necessary for optimal hyperparameter optimization. Future work could explore controlled transparency mechanisms—periodic
  strategy disclosure after competitive advantage windows or differential privacy for configuration sharing—but maintaining some degree of strategic opacity appears
  crucial for driving continuous innovation.

\section{Conclusion}

Our findings reveal that hyperparameter optimisation contains unexploited potential that centralised search strategies often miss. Through competitive economic coordination, Gradients discovers superior fine-tuning configurations across diverse foundation model architectures and task types, achieving 82.8\% win rate against HuggingFace AutoTrain and perfect performance against other leading platforms. The 30-40\% improvements on complex reasoning tasks and 23\% gains for diffusion models indicate that current AutoML approaches explore only a fraction of viable hyperparameter space.


\bibliographystyle{ACM-Reference-Format}
\bibliography{references}

\end{document}